# SEMRes-DDPM: Residual Network Based Diffusion Modelling Applied to Imbalanced Data

Ming Zheng, *Member, IEEE,* Yang Yang, Zhi-Hang Zhao,
Shan-Chao Gan, Yang Chen, Si-Kai Ni and Yang Lu, *Member, IEEE*

***Abstract*—In the field of data mining and machine learning, commonly used classification models cannot effectively learn in unbalanced data. In order to balance the data distribution before model training, oversampling methods are often used to generate data for a small number of classes to solve the problem of classifying unbalanced data. Most of the classical oversampling methods are based on the SMOTE technique, which only focuses on the local information of the data, and therefore the generated data may have the problem of not being realistic enough. In the current oversampling methods based on generative networks, the methods based on GANs can capture the true distribution of data, but there is the problem of pattern collapse and training instability in training; in the oversampling methods based on denoising diffusion probability models, the neural network of the inverse diffusion process using the U-Net is not applicable to tabular data, and although the MLP can be used to replace the U-Net, the problem exists due to the simplicity of the structure and the poor effect of removing noise. problem of poor noise removal. In order to overcome the above problems, we propose a novel oversampling method SEMRes-DDPM.In the SEMRes-DDPM backward diffusion process, a new neural network structure SEMST-ResNet is used, which is suitable for tabular data and has good noise removal effect, and it can generate tabular data with higher quality. Experiments show that the SEMResNet network removes noise better than MLP; SEMRes-DDPM generates data distributions that are closer to the real data distributions than TabDDPM with CWGAN-GP; on 20 real unbalanced tabular datasets with 9 classification models, SEMRes-DDPM improves the quality of the generated tabular data in terms of three evaluation metrics (F1, G-mean, AUC) with better classification performance than other SOTA oversampling methods.**

***Index Terms*—Imbalanced data, Oversampling approach, DDPM**

## I. INTRODUCTION

IN the field of machine learning and data mining, tabular data --- also known as structured data, and unbalanced data distribution is a common phenomenon in that type of data. Data distribution imbalance generally refers to the unbalanced distribution of samples of different categories in a dataset. In a dichotomous or multicategorical dataset affected by data distribution imbalance, categories with fewer samples are referred to as positive or minority categories, while categories with more samples are referred to as negative or majority categories. In actual real-world data, although non-tabular data exists in abundance, tabular data still exists and the analysis of this tabular data is critical for industries such as industry, banks and hospitals that use relational databases. Text tagging[1], Wind Turbine Blade Icing Detection[2], text categorisation[3], fraud detection and disease diagnosis based on medical data [4, 5], to name a few, suffer from the problem of unbalanced distribution of tabulated data. Since traditional classification models are built on relatively balanced data distributions, these algorithms may suffer from different degrees of defects and thus become inefficient in the face of the data distribution imbalance problem. Therefore, improving the performance of classification models in the presence of data distribution imbalance is an important issue that needs to be addressed urgently.

An effective solution to the problem of classifying unbalanced data is through data-level oversampling methods, i.e. synthesising a small number of class samples and adding them to the original unbalanced dataset to bring it into balance [6]. And the current oversampling methods are divided into three main categories: oversampling methods based on statistical learning[6-8], Generative adversarial network based oversampling methods[9-11] and diffusion model based oversampling methods[12, 13]. Statistical learning based oversampling methods are mostly based on SMOTE (Synthetic Minority Over-sampling Technique)[6]. The samples are generated by random interpolation mainly along the line segments connecting the few classes of samples. However, since these methods focus more on the local information of the unbalanced data, the data they generate may not be realistic enough, leading to the occurrence of overfitting. Whereas, oversampling methods based on generative adversarial networks are typically represented by methods based on GAN (generative adversarial networks) and its variants[9-11]. These methods can capture the true distribution of unbalanced data and thus directly generate synthetic minority class samples that are close to the true data distribution. However, it has been shown in the literature that these methods suffer from problems such as difficulty in training and pattern collapse [9].

DDPM(Denoising diffusion probability model)[14] have recently gained a lot of attention and research in generative networks because they tend to outperform other methods in terms of the realism and diversity of the synthetic samples and they are largely free from the problem of training instability and pattern collapse.The most impressive successes of DDPMs are In the field of natural images, it has great advantages in image generation, colouring, segmentation, and

Authors, organization, E-mail…

semantic editing[14-17]. DDPM is also used in the direction of natural language processing, waveform signal processing and time series[18-20], which confirms the universality of DDPM for various types of problems. Recently, Akim Kotelnikov et al. proposed TabDDPM[12], giving the generative network-based oversampling approach another direction to explore in the study of unbalanced tabulated data problems. However, in TabDDPM, the neural network structure U-Net[21] , which is used to remove noise in the backward diffusion process of the diffusion model, is simply replaced with MLP (Multilayer Perceptron)[22]. Therefore, based on the characteristics of the MLP, it is highly likely that the TabDDPM has poor noise removal and incomplete noise removal.

To this end, this paper proposes a new oversampling method SEMRes-DDPM for tabularised data based on denoising diffusion probabilistic model for capturing the distribution of real data, synthesising minority class samples and using it in unbalanced tabularised data classification problems. The method uses a novel hybrid neural network structure for denoising tabularised data in the inverse denoising process. It can effectively remove the noise and extract the features of the original real data, which makes the minority class samples generated by SEMRes-DDPM closer to the real data distribution.

The main contributions of this study to the problem of unbalanced tabular data classification can be summarised as follows:

1) Unlike traditional oversampling methods based on statistical learning that focus only on the local information of the imbalance data, the proposed SEMRes-DDPM focuses more on the global information of the imbalance data, thus learning a more realistic data distribution.
2) Compared with oversampling methods based on generative adversarial networks, the proposed SEMRes-DDPM avoids problems such as training instability and pattern collapse while being able to learn the true distribution of the original unbalanced data.
3) The proposed ResNetBU-TabDDPM oversampling method for noise removal in the backward diffusion process is a specialised denoising network for tabulated data, which removes noise effectively. The problem of unsatisfactory denoising effect of using MLP in TabDDPM is avoided.

The rest of the paper is organised as follows. Section II discusses the related work. Section III describes the proposed method in detail. Section IV describes the experimental methodology used to validate the proposed model, the performance evaluation process, and the classification model evaluation metrics. Section V gives the test experiments and discusses the shortcomings of our proposed method.

## II. Related Work

In this section, work related to the thesis research is reviewed. It mainly covers various oversampling methods oriented towards the problem of unbalanced tabular data classification.

### *A. An oversampling approach towards unbalanced tabular data classification*

Oversampling methods oriented towards unbalanced tabular data classification mainly include three types, (1) oversampling methods based on statistical learning, (2) oversampling methods based on generative adversarial networks, and (3) oversampling methods based on diffusion models.

*1） Statistical learning based oversampling methods:* Among the traditional oversampling methods, the most classic is the SMOTE oversampling method, whose core principle is generated by random interpolation based on the minority class samples that are close neighbours of each other. Among them, Borderline-SMOTE[7] , an improved method of SMOTE, classifies the minority class samples among the samples that are close neighbours of each other, and considers only the minority class samples at the boundary to generate samples that better distinguish the boundary between the samples. And ADASYN (Adaptive Oversampling)[8] assigns different weights to different minority class samples to generate different numbers of samples, which is similar in nature to the SMOTE algorithm. Nowadays, there are also many novel oversampling algorithms, such as GDO[23] uses Gaussian distribution to generate the data, the structure does not need to consider the anomalous data distribution. RBO[24] uses radial basis function to find the suitable data from the minority class to synthesise the data. OREM[25] interpolates the synthesised data by dividing the minority class into clean regions. However, all of the above oversampling methods are based on unbalanced data local information, and there may be a possibility that the generated minority class samples are not real enough, making the classification performance of the classification model on balanced tabular data less than ideal.

*2） An oversampling method based on generative adversarial networks:* GANs are currently popular data synthesis models based on generative networks.GANs generate samples by means of a generator against a discriminator. In fact, in unconstrained generative models, there is no control over the type of data generated, so CGAN[10] generates samples by adding additional conditional information to constrain the model. In order to address the situation where there is pattern collapse in GAN where the KL/JS scatter is used as a loss function, WGAN[9] was proposed that uses Wasserstein distance instead of JS scatter as a loss function to push the probability distribution between the generated data and the real data to the same distribution. Among the many variants of GANs, CWGAN-GP[11] is one of the most complex models, adding both additional conditional information to constrain the model and replacing the JS scatter with the Wasserstein distance. Although data synthesis models based on GANs are able to learn the true distribution of data, they still face problems such as difficult training and pattern collapse.

*3）Diffusion model-based oversampling methods:* The diffusion model is an example of a generative model that is meant to approximate the target distribution through the endpoints of a Markov chain. The Markov chain in the model starts with a given parameter, usually a standard Gaussian distribution. The Gaussian noisy data is obtained by first gradually adding noisy data to the original data through a forward diffusion process. In the backward denoising process, the new noise-free data is obtained by gradually removing the noise through the neural network learning effectively at each step with the application of a known Gaussian kernel. The forward noise addition process and the reverse denoising process are shown in Fig. 1, where $X_t$ is the data after each step of noise addition or denoising. Therefore, in the reverse denoising process, the parameters of the neural network need to be learned by optimising the variational lower bound. More similar in structure to the diffusion model is the denoising score model based on the annealed Langzhi dynamics[13], the denoising score model also has a forward process and a reverse process, and the model uses a score model that learns the gradient of the data distribution for denoising. The data generated by this model has high quality, and some scholars have applied it to the generation of tabular data[26]. Accordingly, the mainstream DDPM (Denoising Diffusion Probabilistic Model) of diffusion models is now more widely used in data synthesis. Among them, TabDDPM is proposed based on the denoising diffusion model probabilistic model and applied to the generation of tabular data. The U-Net network structure in the original denoising diffusion probabilistic model is suitable for image data, but not for tabular data. Therefore, in TabDDPM the model replaces the U-Net network with a simpler MLP. although TabDDPM can generate higher quality data, the simple MLP network may have the problem of undesirable separation of noise and real data during the process of inverse denoising, which may cause the distribution of the generated data to deviate from the distribution of the real data and lead to generation of The quality of the tabulated data is not satisfactory.

## III. Proposed Method

In order to alleviate the limitations of the above mentioned tabular data oriented oversampling methods, we propose SEMRes-DDPM.In this section, the paper will detail the design of SEMRes-DDPM and the main components of the internal denoising network.

### *A. Hybrid Neural Network For Predicting Noise*

Most of the oversampling methods based on statistical learning focus on the local information of unbalanced tabularised data. Although GANs-based oversampling methods are able to learn the global information of unbalanced tabularised data, they still suffer from problems such as difficult training and pattern collapse. Although the DDPM-based TabDDPM oversampling method alleviates the above problems to a certain extent, it simply replaces the U-Net in the DDPM with the MLP, which may suffer from unsatisfactory de-noising effect in the process of backpropagation. As can be seen from Fig. 2, in a scatter plot composed of a set of 2D features, the data after using MLP to remove noise when the number of steps in the backward diffusion process is set at 1000 steps is not very close to the original data, and the original data and the noise data are not completely separated. Therefore, the data obtained from the inverse diffusion process in TabDDPM using MLP for noise removal may not be very satisfactory.

To this end, we propose a hybrid neural network applied to predict noise to alleviate the problem. It consists of several base network modules and is completed by a deep network-like structure. The basic network modules and the overall network framework structure are shown in Fig. 3.

The specific network framework is shown in part a of Fig. 3. Firstly, the final data is obtained through a fully connected layer, then through several base network sub-modules, and finally through a fully connected layer. Where the base network sub-modules are shown in part b of Fig. 3. Where the input is assumed to be X. The first step consists of a residual layer divided into a direct mapping, which is X, and an indirect mapping, which consists mainly of a new multi-attention mechanism with soft thresholding. Then this residual module is defined as shown in (1):

$$SEMST\text{-}ResNet\left(X\right)=H\left(F\left(X\right),W\right)+X \tag{1}$$

Where F is the indirect mapping network module. In the indirect mapping module, the first step is to set up a fully connected layer with an activation function of ReLU to extract and normalise the features as shown in (2):

$$FC\left(X\right)=ReLU\left(BN\left(Linear\left(X\right)\right)\right) \tag{2}$$

Where the function Linear denotes the linear fully connected layer, the prescribed function BN denotes the BorderlineNorm normalised layer and the function ReLU is the ReLU activation function. The second step sets a multi-head self-attention to deal with the correlation between the data, and establishes a connection between the data, which can better allow the data distribution to be reflected, thus facilitating the removal of noise. At the same time, a soft threshold is set at the end of the multi-head self-attention to strengthen the effect of denoising, and the threshold can be set by the multi-layer perceptron automatic learning. And through the Sigmoid activation function ensures that the threshold value of the soft threshold function is positive and within the appropriate range, avoiding the situation where the output is all zero. At the same time, the setting here allows each sample to have its own unique set of thresholds, making the residual network suitable for situations where the noise content of individual samples is different. The layer is defined as shown in (3) and (4):

$$\begin{aligned}&Multi\text{-}Head\ Self\text{-}Attention\left(X\right)=\\&\left\{Self\text{-}Attention\left(Q\left(XW^{Q}\right),K\left(XW^{K}\right),V\left(XW^{v}\right)\right)\right\}\end{aligned} \tag{3}$$

$$Soft\text{-}Thresholding\left(X\right)=Sigmoid\left(BN\left(Linear\left(X\right)\right)\right) \tag{4}$$

n this case, the specific implementation of the mechanism of multiple self-attention is shown in part c of Fig. 3, where *Q*, *K*, and *V* stand for Query, Key and Value, which are varied from the input matrix X, i.e., $XW^Q$、$XW^K$、$XW^V$ , respectively. where $W^Q$, $W^K$, and $W^V$ are the matrices of the training parameters that are multiplied by the dots with *X*, respectively. The overall idea adopts a similar approach to Squeeze-and-Excitation Network attention mechanism with residual contraction network[27] . Then the result obtained by soft thresholding is shown in (5):

$$\tilde{X} = Multi\text{-}Head\ Self\text{-}Attention(X) - Soft\text{-}Thresholding(X) \tag{5}$$

Thus in the indirect mapping module, the combined (2) and (5) can be written in the form of (6)

$$F(X) = Multi\text{-}Head\ Self\text{-}Attention(FC(X)) - Soft\text{-}Thresholding(FC(X)) \tag{6}$$

The above is the base network module, which is repeated as a sub-module many times in the deep network framework. The final form of the overall network structure of the ground is shown in (7):

$$\hat{X} = FC(SEMST\text{-}ResNet(SEMST\text{-}ResNet(FC(X)))) \tag{7}$$

The above is the overall network structure, framework and sub-modules, etc., which can better identify the data distribution in the form of memories and identify the noise through soft thresholding to achieve better denoising.

### *B. SEMRes-DDPM For Tabular Data*

In SEMRes-DDPM, for the tabular dataset $s$={$s_1$, $s_2$…$s_n$}, which consists of *n* data quantities, there are *num* features in each data $s_i$. The character-based data in the data examples are first thermally encoded and processed into numerical data. Then all numerical data are max-min normalised to get the final data $s^{norm}$ that can be used as model input.

In the forward noise addition process of the model, noise is added to each data $s^{norm}$, i.e., the initial input to the model is defined here as $S_0=s^{norm}$, then $q(S_0)$ is the true distribution of the data, from which the data can be obtained by sampling. At this point, the forward diffusion process is defined to add Gaussian noise from $S_t$ to $S_{t-1}$ at each time step, which is used for the learning process through a probability distribution form $q(S_t|S_{t-1})$. Under the Markov chain, the data distribution $S_t$ at any moment can be obtained through the joint probability density distribution, as shown in (8):

$$q(S_t \mid S_0) = N\left(S_t; \sqrt{\bar{\alpha}_t} S_0, (1-\bar{\alpha}_t) I\right) \tag{8}$$

Where $\bar{\alpha}_t = \alpha_t \alpha_{t-1} \ldots \alpha_0$, and $\alpha_t$=1-$\beta_t$, assuming that $\beta_t$ is the variance of the added noise, also known as the diffusivity. And when the number of added noise steps T is large enough, $\bar{\alpha}_t$ tends to 0 and $1-\bar{\alpha}_t$ tens to 1, then $S_t$ tends to the standard Gaussian distribution.

After forward noise addition, the $S_T$ is gradually denoised in the reverse process to restore it to $S_0$. The reverse process is still a Markov chain, however, the reverse process $q(S_{t-1}| S_t)$ is unlikely to be derived directly from the theory, and is therefore fitted by the neural network (8). At this point the inverse denoising process is defined as shown in (9):

$$p_\theta(S_{t-1} \mid S_t) = N\left(S_{t-1}; \mu_\theta(S_t, t), \Sigma_\theta(S_t, t)\right) \tag{9}$$

Where $\theta$ is the parameter of the neural network, $\mu_\theta$, $\Sigma_\theta$are the mean and variance of the contained parameters. During the optimisation training process, the objective of the model is to obtain the most realistic $S_0$, possible, i.e., to solve for the parameters of the model, $\theta$. In SEMRes-DDPM, the optimisation objective is transformed into minimising the KL scatter between the posterior distribution and the Gaussian distribution parameterised by the network by means of the Bayesian formulation and combining it with the construction of the Gaussian function. Thus the final optimisation objective of SEMRes-DDPM is obtained defined as shown in (10):

$$L_{t-1}^{simple} = \mathrm{E}_{S_0, \bar{z}_t \sim N(0,I)}\left[\left\| \bar{z}_t - z_\theta\left(\sqrt{\bar{\alpha}_t} S_0 + \sqrt{1-\bar{\alpha}_t}\, \bar{z}_t, t\right)\right\|^2\right] \tag{10}$$

Where $z_\theta$ denotes the fitting term of the neural network. Finally, after training the model through the loss function constructed by the optimisation objective, the parameters are brought into (5) and put into the pure noise data through a step-by-step denoising to get the generated tabulated data.

## IV. Experient And Results

In this section the paper describes the experimental design and results and details an extensive evaluation of the proposed model SEMRes-DDPM according to existing schemes.

### *A. Data Set And Environmental Configuration*

*1）Real Imbalanced Datasets:* In order to systematically investigate the properties of the generated data, the paper considers 20 real-world datasets derived from the public dataset Keel.The 20 datasets have different sizes, properties, imbalance ratios, number of features, and distributions. The datasets have been previously used in the evaluation of various tabular data models[23, 24]. The dataset and its complete list is shown in Table I.

*2）Environmental Configuration:* For this experimental environment, experiments were conducted under Python and the environment was configured using Conda. Among the classical statistical methods experiments were conducted using the unbalanced data package Imblearn calling various methods. Generative Adversarial Networks in Deep Learning Networks were experimented using the Keras framework which is now integrated in Tensorflow. Diffusion models in deep learning were experimented with the Pytorch framework. The specific versions of the calling libraries are shown in Table II, and CUDA11.7 was chosen as the experimental environment for GPUs, and the CUDNN acceleration module was used.

### *B. Research Method And Evaluation Metrics*

*1）Remove Noise Comparison:* n the comparison of noise removal from neural networks, raw data will be extracted from the training of the diffusion model, denoised data and compared. Firstly a qualitative comparative analysis, where the denoised data as well as the extracted noise data will be directly compared and presented through visualisation in the experiment. Then there is a quantitative comparative analysis, where the final performance results will be obtained by comparing the PSNR peak signal-to-noise ratio, an important metric for denoising evaluation, in the experiment.

*2）Qualitative Comparison:* In the qualitative comparison of the models, the experiment will compare the ability of different models to model the distribution of data by comparing the distribution of the generated data and the real data of different models and displaying them through visualisation to reflect the similarity between the distribution of the generated data and the real data. Not only comparing the distribution of generated data, but also calculating the correlation coefficients between real and synthetic data from different datasets will be used to reflect the differences between the data. In order to calculate the correlation coefficient, the experiment uses Pearson's correlation coefficient [28] to represent the numerical correlation.

*3）Machine Learning Efficiency:* In the evaluation of machine learning efficiency, experiments are conducted to compare classification performance using different classification models on unbalanced tabulated data. The generalisation ability of the models is evaluated after performing effective experimental methods. The original dataset is divided into training and test sets, and a few classes from the training set are recombined into a balanced training dataset for use in the classification model after an oversampling method. The classification model is trained using the new balanced dataset and then tested on the divided test set to get the test results. Nine classification models are used in the experiments, namely, simple Bayesian classification model based on Gaussian distribution, simple Bayesian classification model based on Bernoulli distribution, K-nearest neighbour classification model, logistic regression classification model, random forest classification model, decision tree classification model, gradient boosting classification model, support vector machine classification model and multilayer perceptron classification model[29-35].

Through the classification results of the classification model on the test set, three evaluation metrics will be used to measure the performance of the classification model, so as to reflect the specific performance of the data synthesis method in terms of machine learning efficiency. Firstly, the confusion matrix of the classification results is obtained, and the three evaluation metrics are F-measure, G-mean and the area under the curve (AUC) of the subject's work characteristics (ROC) using the confusion matrix[36]. F-measure usually responds to the average reconciliation metrics of checking accuracy and checking rate, and then combines them to respond to the effect of the classification model.G-mean usually responds to the average reconciliation metrics of checking accuracy and checking rate of negative cases, and then combines them to respond to the effect of the classification model.G-mean usually responds to the average reconciliation metrics of checking accuracy and checking rate of negative cases. G-mean usually responds to the average reconciliation index of the positive and negative case checking rate. G-mean is more suitable for the evaluation of classification on unbalanced datasets, which can assess the performance of the model more comprehensively.The area enclosed by the ROC curve and the coordinate axis is the AUC area value. Since the ROC curve depends on the classification threshold, the AUC is a useful metric to evaluate the performance of classification models because it is independent of the decision criterion[37].

### *C. Experimental Process*

For comparing the machine learning efficiency, traditional oversampling methods SMOTE, ADASYN, advanced oversampling methods RBO, GDO and OREM were selected for the experiments.Similarly, GAN-based oversampling methods CGAN, CWGAN-GP were also selected, and on the same kind of diffusion model and the most recent TabDDPM was selected.The data synthesised by the above seven data synthesis methods plus raw data are compared with the methods proposed in the paper. In the qualitative comparison, the experiment will be compared with the data synthesised by TabDPPM with fixed parameters. In the denoising effect experiments, the focus is on comparing with the data obtained by denoising the MLP network used in TabDDPM.

*2）Operation Process：* The specific operation flow of the experiment is presented in Fig. 4. The process is mainly to verify the machine learning effect, firstly, the dataset is divided into ten copies as $\{D_1...D_{10}\}$, so as to carry out ten-fold cross-validation. One of the copies of the divided dataset is selected sequentially as the test set, and the remaining nine copies are used as the training set. Then the minority class samples in the training set are extracted, and the data synthesis method is applied to achieve balance with the majority class data after oversampling, and combined as the new balanced training set. The new balanced training set is applied to the classification model for training, and then the trained classification model is tested on the test set to get the performance evaluation results. A total of ten experiments are conducted and the average and standard deviation of the ten results are taken as the final results. The denoising data for the denoising experiments and the original data will be obtained from the training of TabDDPM and SEMRes-DDPM, and the denoising step size of the two models will be fixed at 1000 steps to ensure that the models are comparable under the same number of denoising steps. The experimental data for qualitative comparison will be obtained from the data generated by TabDDPM, SEMRes-DDPM and CWGAN-GP, where the denoising step size of the TabDDPM, SEMRes-DDPM models is fixed at 1000 steps, and the parameters of CWGAN-GP are kept as the original paper parameters.

### *D. Experimental results and analysis*

*1）Remove Noise Comparison Visualization And Results:* Fig. 5 shows the scatter plots constructed by SEMST-ResNet, MLP denoising the data under the same number of steps and the original data with two sets of feature data extracted from each. Among them, the scatterplot composed of data obtained by SEMST-ResNet denoising is obviously closer to the original data, while the scatterplot composed of data obtained by MLP denoising is less effective. Therefore, with the same number of denoising steps, the effect of SEMST-ResNet denoising is significantly better than that of MLP.

In our experiments, we also calculated the peak signal-to-noise ratios of a complete set of data and the two feature data in that data after noise addition, the data obtained after denoising by MLP and SEMST-ResNet, and the original data. The results obtained in Table III show that the peak SNR of both SEMST-ResNet and the original data is greater than that of MLP, which also means that the data obtained from SEMST-ResNet denoising is of higher quality. Therefore, with the same number of denoising steps, the data quality obtained from SEMST-ResNet denoising is better than MLP.

*2）Qualitative Comparison Visualization:* In the experiment, we fixed the number of training iterations in the denoising diffusion probability model to 20000, and the number of denoising timesteps to 1000. with the same two parameters, the data generated by the training of TabDDPM, SEMRes-DDPM, and CWGAN-GP were compared with the original data through the visualisation of the distributions, and the results are shown in Fig. 6, where it can be seen that the distributions of the data generated by SEMRes-DDPM are mostly closer to the real data in terms of probability density distributions and cumulative probability distributions than those generated by TabDDPM and CWGAN-GP. probability density distribution and cumulative probability distribution, under the same denoising time step, the distribution of data generated by SEMRes-DDPM is mostly closer to the distribution of the real data than that of TabDDPM and CWGAN-GP.Therefore, from the closeness of the final distributions, it can be seen that the distribution of data generated by SEMRes-DDPM is also more realistic than that of TabDDPM and CWGAN -GP is more realistic. And with the same denoising time step, our proposed hybrid neural network denoises better than MLP.

The experiments were conducted by generating data from CWGAN-GP, TabDDPM and SEMRes-DDPM and the Pearson's correlation correlation coefficients of the generated data with the original data were calculated. The absolute values of the correlation coefficients calculated from the real data and two of the feature data sets are given in Table IV. Most of the correlation coefficients of SEMRes-DDPM with the real data are larger than those of CWGAN-GP and TabDDPM; therefore, the data generated by SEMRes-DDPM are closer to the real data than the data generated by CWGAN-GP and TabDDPM.

*3）Machine Learning Efficiency Results And Analysis:* 由 From Table V, it can be concluded that among the average efficiencies of the 20 datasets with 9 classification models, the average efficiencies of the models obtained after training the classification models on the balanced data after the SEMRes-DDPM oversampling in most cases are mostly better than the other oversampling methods for the three evaluation metrics, F1, G-mean, and AUC, under the test set. Therefore, the oversampling method based on SEMRes-DDPM model performs better in machine learning efficiency compared to other oversampling methods.

## V. Disscussion

In order to verify the validity and practical performance of the proposed model, we have also conducted the Friexdman test based on algorithm ranking for the average efficiency of the classifiers of all the algorithms on all the datasets. In Friexdman test, the scoring metrics used are F1, G-mean and AUC and the evaluation metrics are the average of the evaluation metrics obtained from the data synthesised by each algorithm in each dataset under multiple classifier tests, Friexdman test is used to determine whether the performance of these methods is the same. If not, the hypothesis that all methods perform equally is rejected, indicating that there is a significant difference in the performance of these methods. Therefore a post hoc test is required to further differentiate between these methods. Finally, a Nemenyi follow-up test is used to check whether there is a significant difference between any two methods.The Nemenyi follow-up test determines the gap between the two methods based on a critical range. If the gap between the mean order values of the two methods exceeds the critical range, the hypothesis that the two methods have the same performance is rejected at the appropriate confidence level.

The results and average rankings obtained by Friexdman test are shown in Table VI, from which it can be seen that not all algorithms perform exactly the same and the p-value of the test obtained by Friexdman test is much less than the significance level, therefore, Nemenyi follow-up test is required, and the final results obtained are shown in Fig. 7, where the vertical axis represents each method and the horizontal axis is the average order value. For each method, the distant point indicates its mean order value, and the horizontal line centred on the origin indicates the size of the critical range. Among them, our method is significantly better than 3-6 methods in each evaluation index.

Although the data generated by our proposed model SEMRes-DDPM achieved better results in machine learning efficiency and qualitative comparison, there are still some shortcomings, in which the model spends a large amount of time for training, and if the training and synthesis of small batches of samples are carried out may not be as good as other models if they are considered from the perspective of both accuracy and time. Therefore, reducing the training time of the model is a problem that needs to be solved, and at the same time, whether the model can add labels to the training process, so that the labels have an impact on the training data, so as to improve the quality of the synthetic data is also worth thinking about.

## VI. Conclusion

Unbalanced data distribution is a common phenomenon in theoretical studies and practical applications. Classification models are mostly based on datasets with relatively balanced category distributions. However, applying unbalanced data to classification models may make them inefficient. To alleviate the problem caused by imbalanced data distribution, oversampling methods have been proposed. Oversampling methods are techniques that generate few classes of data to balance the original dataset. Although these oversampling methods have been shown to improve performance, traditional oversampling methods based on SMOTE and its variants focus on the local information of the data, and therefore generate less realistic data. GANs-based oversampling methods often encounter problems of pattern collapse and training instability. In addition, TabDDPM oversampling methods suffer from poor noise removal in the inverse denoising process. To address these problems, we propose a novel DDPM-based oversampling method, SEMRes-DDPM, which employs a new network module, SEMST-ResNet, suitable for denoising tabular data.

The new network module SEMST-ResNet for tabular data denoising is significantly better than MLP in terms of denoising effect, thus ensuring that SEMRes-DDPM generates more realistic data. In comparison with CWGAN-GP and TabDDPM, SEMRes-DDPM generates data distributions that are closer to the original data distribution. In terms of machine learning efficiency, the original unbalanced data is balanced by the few classes of data generated by SEMRes-DDPM and used in the training of the classification model, which makes the classification performance of the classification model on the test set become better and improves the generalisation performance of the classification model, and all of them outperform the other oversampling methods. In our future research, we intend to apply the SEMRes-DDPM oversampling method to real-world imbalanced tabular data classification problems related to specific practical applications.

## ACKNOWLEDGMENT

This work was supported in part by the This work was supported by the supported by National Natural Science Foundation of China (62306009, 62272006); the Major Project of Natural Science Research in Colleges and Universities of Anhui Province (KJ2021ZD0007); Wuhu Science and Technology Bureau Project (2022jc11).

FIGURE

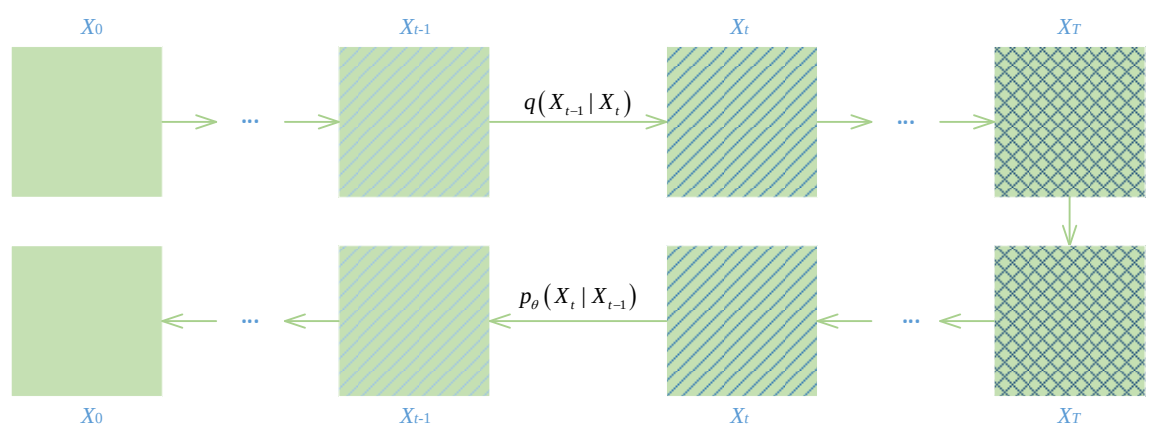

Fig. 1. Forward noise addition and reverse denoising process diagrams

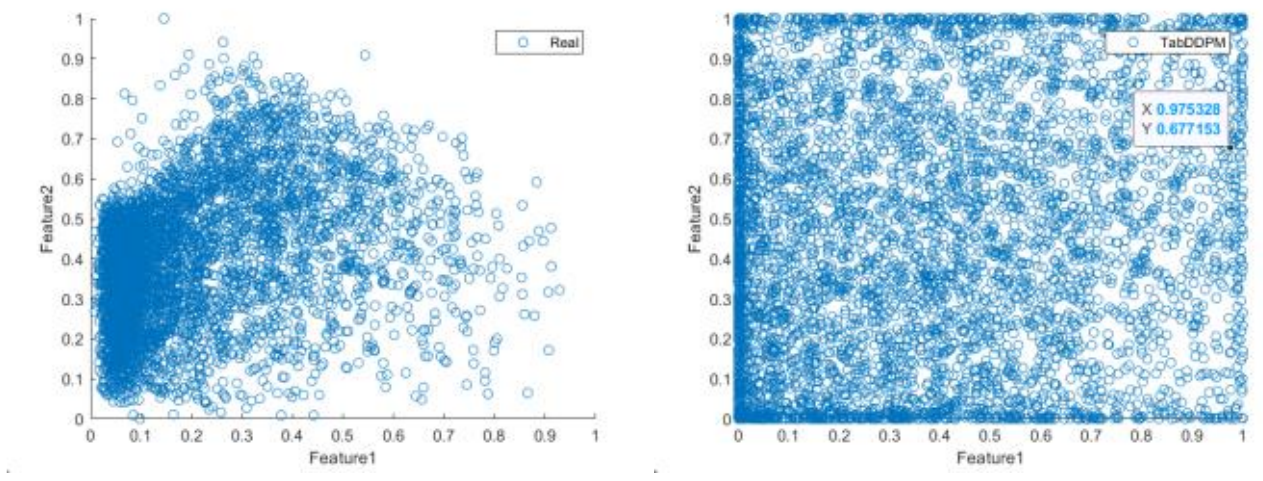

Fig. 2. Scatterplot of the data obtained from MLP denoising in TabDDPM with two sets of features of the original data

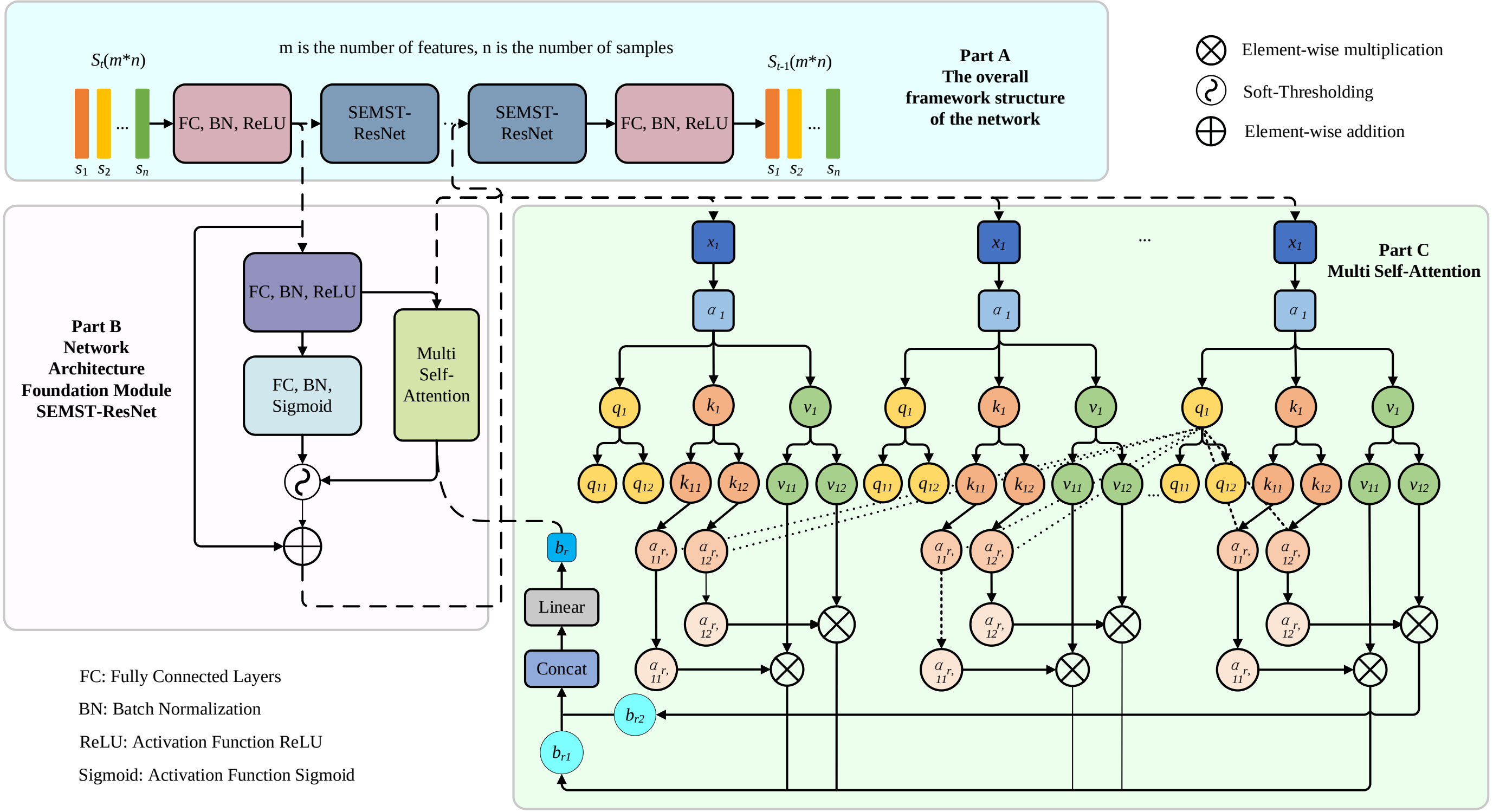

Fig. 3. Overall framework and sub-modules of the hybrid neural network (part a is the overall framework, part b is the SEMST-ResNet module and part c is the multi-head self-attention mechanism)

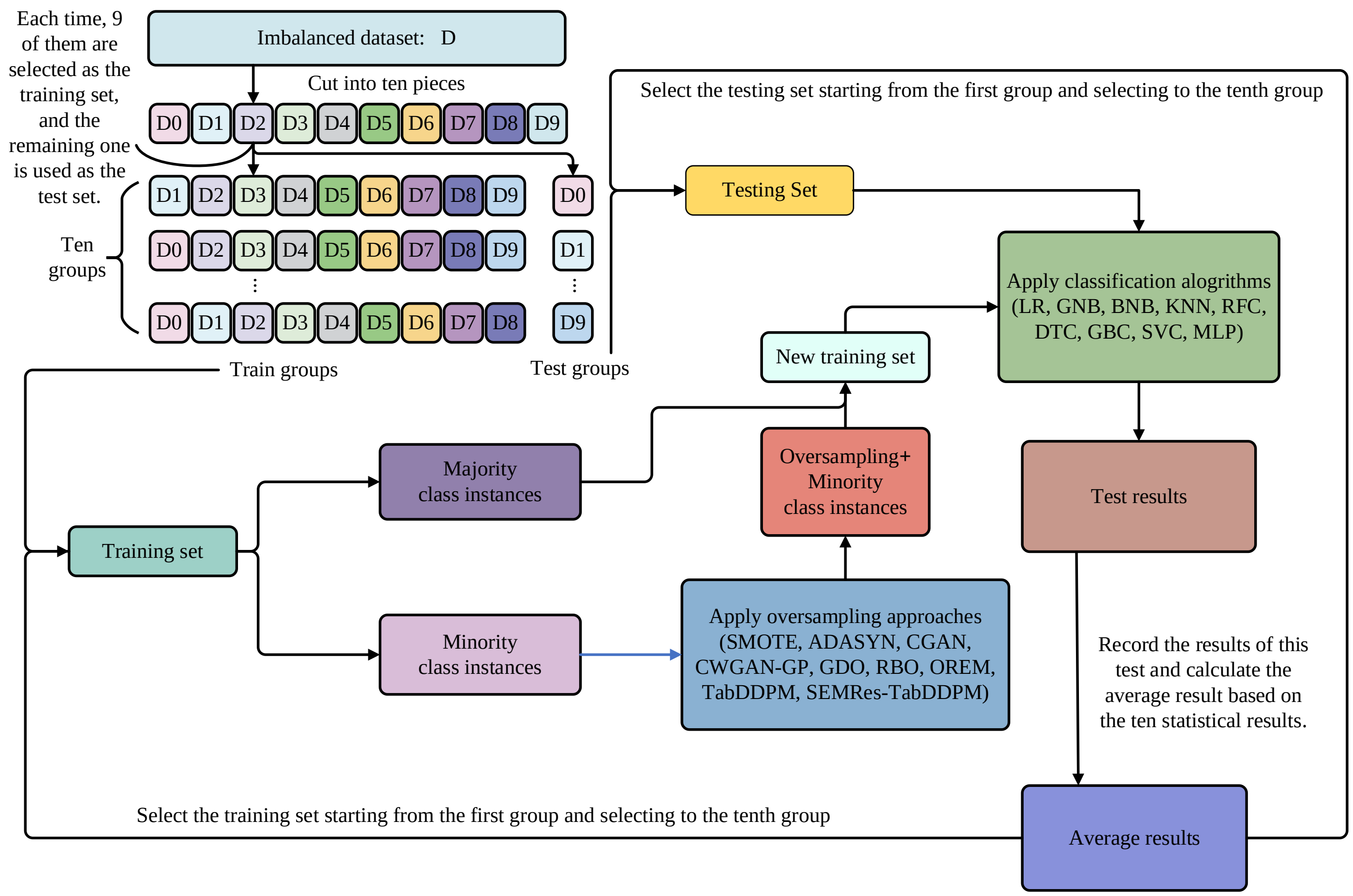

Fig.4. Flowchart of the experiment

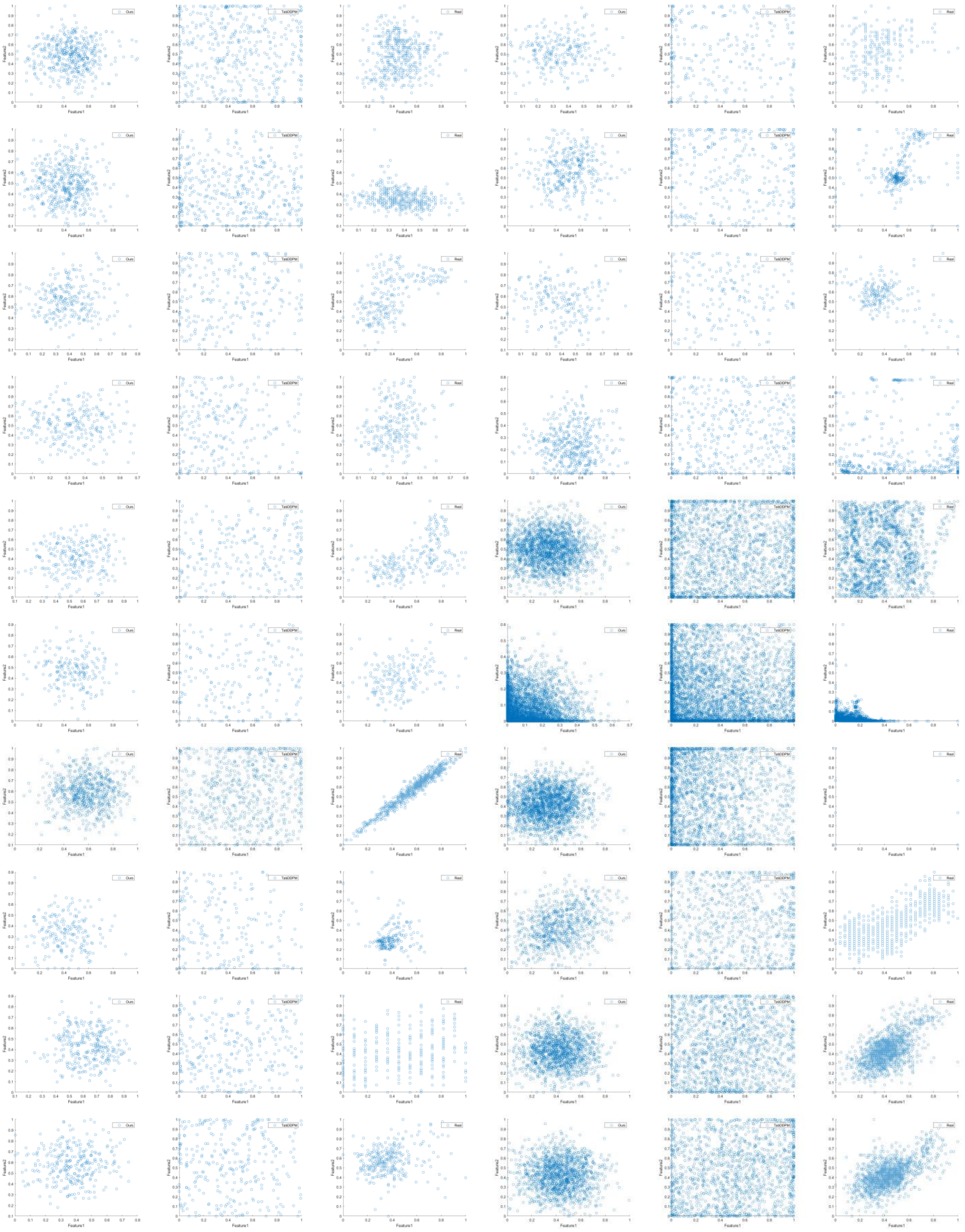

Fig. 5. 20 sets of datasets under MLP, our method removes noise from the data with the original data constituting a 2D scatter

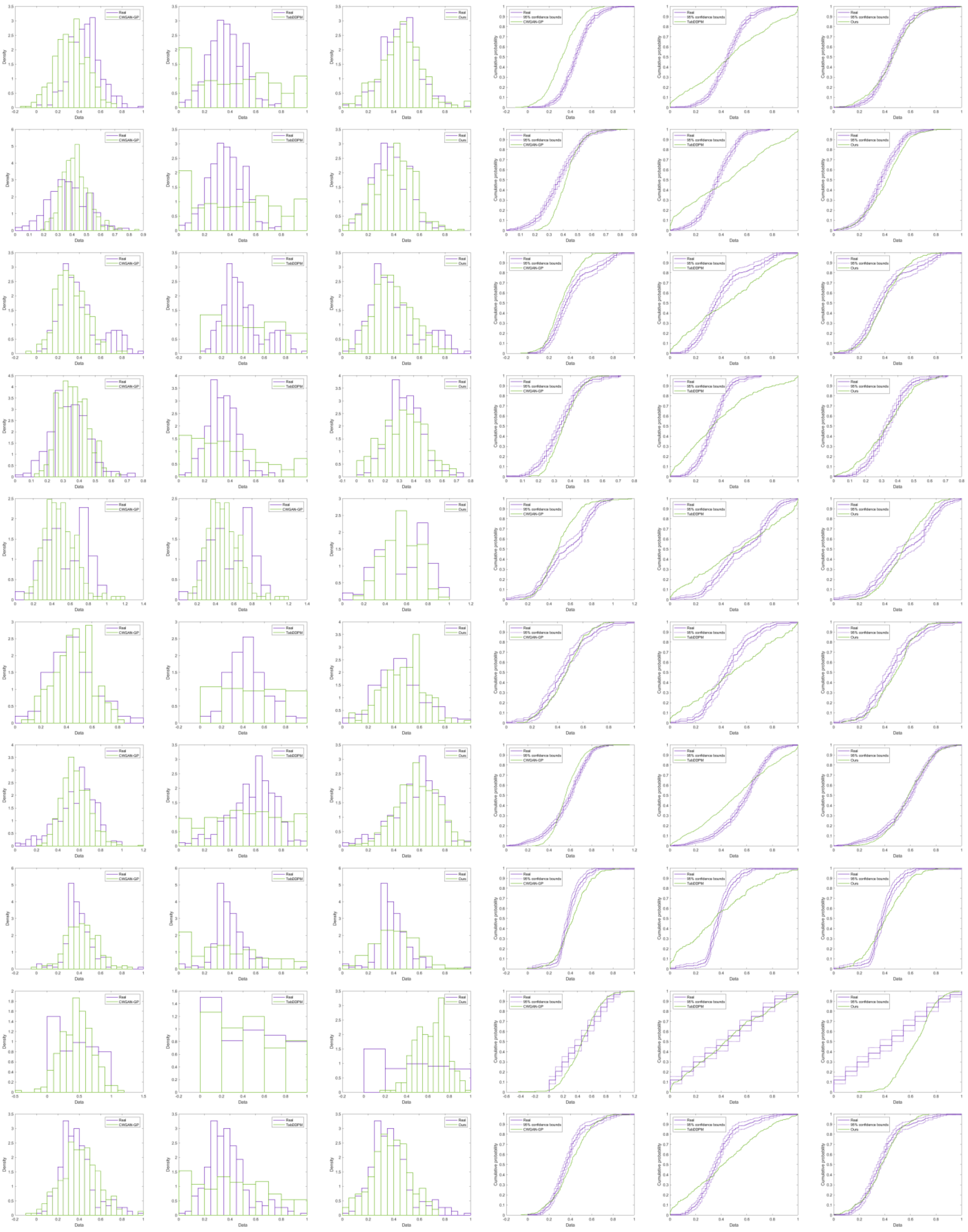

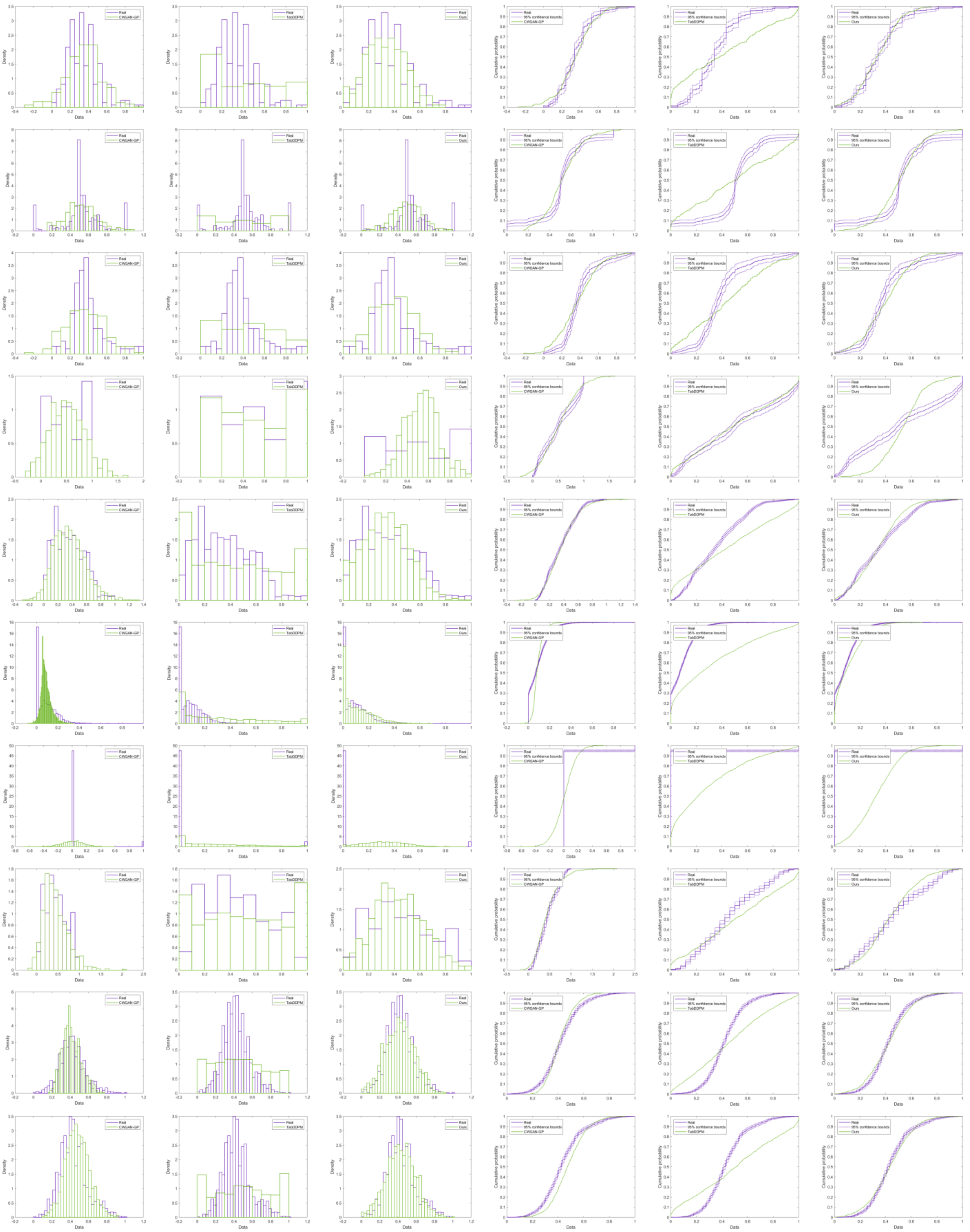

Fig. 6. Density distribution of data generated by CWGAN-GP, TabDDPM and our method under 20 sets of datasets fitted to the original data versus cumulative density distributions

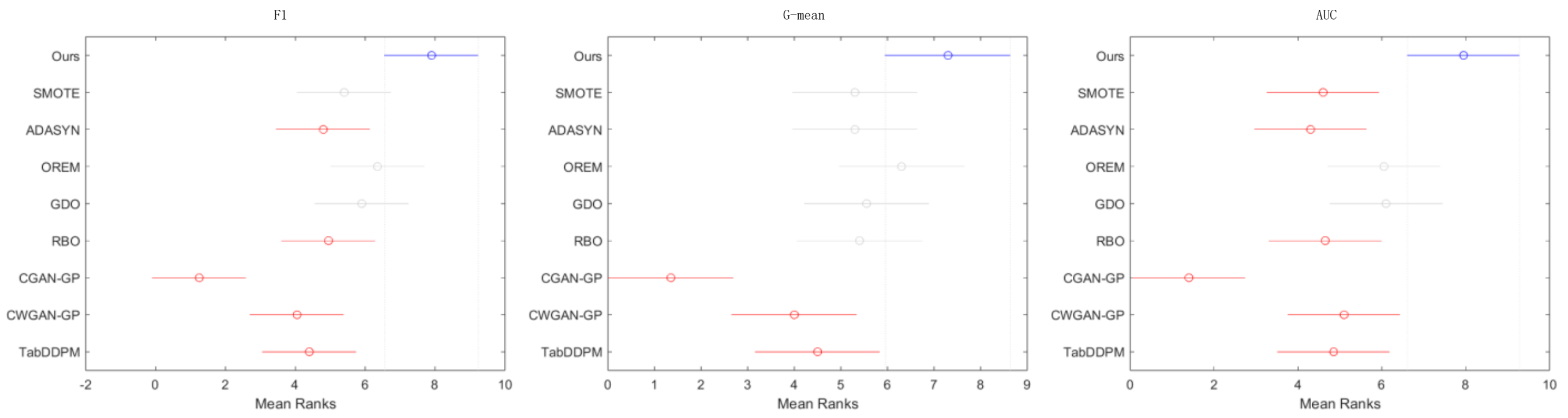

Fig.7. Subsequent inspection plots for multiple oversampling algorithms

TABLE

TABLE I

Details of the datasets used in the assessment

| ID | Data sets | Features | Number of minority | Number of majority | IR |
|---|---|---|---|---|---|
| abalone9-18 | 731 | 8 | 42 | 689 | 16.4 |
| ecoli4 | 336 | 7 | 20 | 316 | 15.8 |
| page-blocks-1-3-vs-4 | 472 | 10 | 28 | 444 | 15.86 |
| yeast5 | 1484 | 8 | 44 | 1440 | 32.73 |
| yeast-0-5-6-7-9-vs-4 | 528 | 8 | 51 | 477 | 9.35 |
| yeast-2-vs-4 | 514 | 9 | 51 | 463 | 9.08 |
| ecoli1 | 336 | 7 | 77 | 259 | 3.36 |
| ecoli2 | 336 | 7 | 52 | 284 | 5.46 |
| ecoli3 | 336 | 7 | 35 | 301 | 8.6 |
| ecoli-0-vs-1 | 220 | 7 | 77 | 143 | 1.86 |
| glass-0-1-2-3-vs-4-5-6 | 214 | 9 | 51 | 163 | 3.2 |
| haberman | 306 | 3 | 81 | 225 | 2.78 |
| newthyroid2 | 215 | 5 | 35 | 180 | 5.14 |
| segment0 | 2308 | 19 | 329 | 1979 | 6.02 |
| vehicle2 | 846 | 18 | 218 | 628 | 2.88 |
| yeast3 | 1484 | 8 | 163 | 1321 | 8.1 |
| heart | 270 | 13 | 120 | 150 | 1.3 |
| ionosphere | 351 | 33 | 126 | 225 | 1.8 |
| spambase | 4597 | 57 | 1812 | 2785 | 1.5 |
| titanic | 2201 | 3 | 711 | 1490 | 1.9 |

TABLE II

Experimental environment version

| Environment/Library | version |
|---|---|
| Python | 3.9.0 |
| Keras | 2.10.0 |
| Pytorch | 2.0.1+GPU |
| Tersorflow | 2.10.0 |
| Imblearn | 0.0 |

TABLE III

Peak signal-to-noise ratio of MLP and SEMST-ResNet denoised data to original data

| | MLP | Ours |
|---|---|---|
| abalone9-18 | 8.6562 | **11.7821** |
| ecoli4 | 8.2982 | **12.4914** |
| page-blocks-1-3-vs-4 | 7.5050 | **12.9346** |
| yeast5 | 8.7605 | **11.9291** |
| yeast-0-5-6-7-9-vs-4 | 8.8140 | **13.6462** |
| yeast-2-vs-4 | 9.4323 | **12.0783** |
| ecoli1 | 8.7375 | **12.8365** |
| ecoli2 | 8.8673 | **12.8910** |
| ecoli3 | 8.7733 | **12.5597** |
| ecoli-0-vs-1 | 9.1937 | **12.2647** |
| glass-0-1-2-3-vs-4-5-6 | 8.6486 | **13.3104** |
| haberman | 9.0254 | **10.7228** |
| newthyroid2 | 9.5799 | **11.7295** |
| segment0 | 7.5942 | **11.9194** |
| vehicle2 | 8.1782 | **11.7086** |
| yeast3 | 9.9409 | **14.5842** |
| heart | 6.2692 | **8.5499** |
| ionosphere | 7.0608 | **9.6786** |
| spambase | 7.7084 | **18.4505** |
| titanic | 6.1010 | **7.2775** |

TABLE IV

orrelation coefficients of CWGAN-GP, TabDDPM, SEMRes-TabDDPM with raw data

| | CWGAN-GP | TabDDPM | Ours |
|---|---|---|---|
| abalone9-18 | **0.8804** | 0.7213 | 0.7548 |
| ecoli4 | 0.6659 | **0.8612** | 0.1953 |
| page-blocks-1-3-vs-4 | 0.8694 | 0.8826 | **0.9823** |
| yeast5 | 0.0304 | 0.5075 | **0.7584** |
| yeast-0-5-6-7-9-vs-4 | 0.2915 | 0.2474 | **0.8718** |
| yeast-2-vs-4 | 0.7593 | 0.5711 | **0.8718** |
| ecoli1 | 0.7154 | 0.6265 | **0.7250** |
| ecoli2 | **0.9217** | 0.1210 | 0.8830 |
| ecoli3 | 0.2250 | **0.4994** | 0.4682 |
| ecoli-0-vs-1 | **0.9718** | 0.0805 | 0.9674 |
| glass-0-1-2-3-vs-4-5-6 | 0.1015 | **0.9061** | 0.3456 |
| haberman | 0.6486 | 0.6611 | **0.6885** |
| newthyroid2 | 0.1120 | 0,7307 | **0.9786** |
| segment0 | 0.1908 | 0.3037 | **0.4553** |
| vehicle2 | 0.1496 | 0.5779 | **0.7671** |
| yeast3 | **0.9027** | 0.3998 | 0.5885 |
| heart | **0.6138** | 0.2056 | 0.5753 |
| ionosphere | 0.4073 | 0.4515 | **0.4797** |
| spambase | 0.2016 | 0.1730 | **0.4555** |
| titanic | 0.7240 | **0.9229** | 0.7979 |

TABLE V

Average efficiency of multiple oversampling algorithms for F1, G-mean, AUC for 9 classifiers under 20 datasets

| F1 | NONE | SMOTE | ADASYN | OREM | GDO | RBO | CGAN-GP | CWGAN-GP | TabDDPM | Ours |
|---|---|---|---|---|---|---|---|---|---|---|
| abalone9-18 | 0.1658 | 0.3059 | 0.3189 | 0.3174 | 0.3295 | 0.3152 | 0.2488 | 0.2890 | 0.2897 | **0.3366** |
| ecoli4 | 0.5881 | 0.7349 | 0.6968 | **0.7445** | 0.7141 | 0.7220 | 0.6134 | 0.6718 | 0.6430 | 0.6569 |
| page-blocks-1-3-vs-4 | 0.3195 | 0.4979 | 0.4882 | 0.5000 | 0.5271 | 0.5026 | 0.3673 | 0.4908 | 0.4969 | **0.5321** |
| yeast5 | 0.2125 | 0.4014 | 0.3811 | 0.4067 | **0.4294** | 0.3842 | 0.3219 | 0.4232 | 0.4057 | 0.4238 |
| yeast-0-5-6-7-9-vs-4 | 0.1485 | 0.2769 | 0.2668 | 0.5984 | **0.6547** | 0.5924 | 0.5026 | 0.6456 | 0.6273 | 0.6183 |
| yeast-2-vs-4 | 0.6254 | 0.7047 | 0.6739 | 0.7010 | 0.6535 | 0.6671 | 0.6234 | 0.6540 | 0.6634 | **0.8300** |
| ecoli1 | 0.5523 | 0.6554 | 0.6160 | 0.6642 | 0.6810 | 0.6465 | 0.5666 | 0.6524 | 0.6687 | **0.8345** |
| ecoli2 | 0.3431 | 0.5194 | 0.5135 | 0.5341 | 0.5483 | 0.4903 | 0.4243 | 0.5039 | 0.5261 | **0.6626** |

| | | | | | | | | | |
|---|---|---|---|---|---|---|---|---|---|
| ecoli3 | 0.8547 | 0.8776 | 0.8902 | 0.8660 | 0.8298 | 0.8484 | 0.8030 | 0.8226 | 0.8299 | **0.9739** |
| ecoli-0-vs-1 | 0.8151 | 0.8403 | 0.8316 | 0.8480 | 0.8417 | 0.8478 | 0.8224 | 0.8328 | 0.8370 | **0.8523** |
| glass-0-1-2-3-vs-4-5-6 | 0.1571 | 0.3785 | 0.4185 | 0.4207 | 0.4100 | 0.3916 | 0.2660 | 0.4154 | 0.3936 | **0.4299** |
| haberman | 0.7154 | 0.8391 | 0.8290 | 0.8290 | **0.8522** | 0.8272 | 0.7257 | 0.8208 | 0.8242 | 0.8320 |
| newthyroid2 | 0.8240 | 0.8552 | 0.8162 | 0.8614 | **0.8691** | 0.8583 | 0.7440 | 0.8388 | 0.7599 | 0.8637 |
| segment0 | 0.7697 | 0.8227 | 0.8243 | 0.8211 | 0.8139 | 0.8236 | 0.6928 | 0.7864 | 0.7630 | **0.8245** |
| vehicle2 | 0.4700 | 0.6134 | 0.5850 | 0.6181 | 0.6378 | 0.6005 | 0.4981 | **0.6424** | 0.6166 | 0.6210 |
| yeast3 | 0.7892 | 0.7822 | 0.7776 | 0.7908 | 0.7883 | 0.7829 | 0.7824 | 0.7891 | 0.7892 | **0.7907** |
| heart | 0.8246 | 0.8434 | **0.8488** | 0.8307 | 0.8302 | 0.8382 | 0.8121 | 0.8317 | 0.8377 | 0.8450 |
| ionosphere | 0.8796 | 0.8855 | 0.8836 | 0.8824 | 0.8600 | 0.8790 | 0.8282 | 0.8612 | 0.8206 | **0.8911** |
| spambase | 0.5956 | 0.6998 | 0.7023 | 0.6986 | 0.6760 | 0.7059 | 0.5598 | 0.6558 | **0.7319** | 0.7228 |
| titanic | 0.5720 | 0.5967 | **0.5983** | 0.5934 | 0.5688 | 0.5770 | 0.5447 | 0.5686 | 0.5755 | 0.5759 |

| G-mean | NONE | SMOTE | ADASYN | OREM | GDO | RBO | CGAN-GP | CWGAN-GP | TabDDPM | Ours |
|---|---|---|---|---|---|---|---|---|---|---|
| abalone9-18 | 0.2569 | 0.6590 | 0.6683 | **0.6903** | 0.6489 | 0.6415 | 0.4466 | 0.5574 | 0.5909 | 0.6566 |
| ecoli4 | 0.6324 | **0.9170** | 0.8685 | 0.9125 | 0.9002 | 0.9065 | 0.6851 | 0.7831 | 0.8440 | 0.8511 |
| page-blocks-1-3-vs-4 | 0.4494 | 0.8130 | 0.8115 | **0.8357** | 0.8170 | 0.7965 | 0.5074 | 0.7832 | 0.8100 | 0.8246 |
| yeast5 | 0.2714 | 0.6075 | 0.6086 | 0.6358 | 0.6505 | 0.6172 | 0.4553 | 0.6428 | 0.6319 | **0.6665** |
| yeast-0-5-6-7-9-vs-4 | 0.2288 | 0.5546 | 0.5510 | 0.7449 | 0.7749 | 0.7564 | 0.5825 | 0.7707 | 0.7695 | **0.7800** |
| yeast-2-vs-4 | 0.6971 | 0.7660 | 0.7510 | 0.7687 | 0.7417 | 0.7627 | 0.7036 | 0.7457 | 0.7532 | **0.9039** |
| ecoli1 | 0.6157 | 0.7625 | 0.7264 | 0.7740 | 0.7935 | 0.7756 | 0.6502 | 0.7685 | 0.7858 | **0.9322** |
| ecoli2 | 0.4449 | 0.7324 | 0.7402 | 0.7457 | 0.7358 | 0.6998 | 0.5422 | 0.7140 | 0.7247 | **0.8894** |
| ecoli3 | 0.8583 | 0.8598 | 0.8516 | 0.8602 | 0.8428 | 0.8554 | 0.8192 | 0.8356 | 0.8403 | **0.9793** |
| ecoli-0-vs-1 | 0.8612 | 0.8989 | **0.9123** | 0.9019 | 0.8989 | 0.9053 | 0.8805 | 0.8960 | 0.8963 | 0.8809 |
| glass-0-1-2-3-vs-4-5-6 | 0.2299 | 0.5038 | 0.5235 | 0.5351 | 0.5440 | 0.5091 | 0.3727 | 0.5463 | 0.5238 | **0.5521** |
| haberman | 0.7543 | 0.8600 | 0.8669 | 0.8528 | 0.8830 | 0.8733 | 0.7663 | 0.8568 | 0.8719 | **0.8847** |
| newthyroid2 | 0.8591 | 0.9092 | 0.8964 | 0.9121 | 0.9240 | 0.9207 | 0.7543 | 0.8667 | 0.8165 | **0.9248** |
| segment0 | 0.8115 | 0.8670 | **0.8728** | 0.8673 | 0.8609 | 0.8651 | 0.7329 | 0.8389 | 0.8227 | 0.8626 |
| vehicle2 | 0.5604 | 0.7583 | 0.7505 | 0.7690 | 0.7851 | 0.7760 | 0.6015 | 0.7810 | 0.7733 | **0.7881** |
| yeast3 | 0.8059 | 0.7986 | 0.7917 | 0.8048 | 0.8036 | 0.7990 | 0.8004 | 0.8060 | 0.8054 | **0.8061** |
| heart | 0.8486 | 0.8682 | **0.8723** | 0.8568 | 0.8549 | 0.8614 | 0.8382 | 0.8563 | 0.8605 | 0.8670 |
| ionosphere | 0.8973 | **0.9044** | 0.9044 | 0.9012 | 0.8805 | 0.8988 | 0.8545 | 0.8814 | 0.8483 | 0.8487 |
| spambase | 0.6550 | 0.7542 | 0.7569 | 0.7530 | 0.7362 | 0.7584 | 0.6228 | 0.7223 | 0.7804 | **0.7914** |
| titanic | 0.6506 | 0.6839 | **0.6887** | 0.6771 | 0.6474 | 0.6593 | 0.6293 | 0.6509 | 0.6569 | 0.6566 |

| AUC | NONE | SMOTE | ADASYN | OREM | GDO | RBO | CGAN-GP | CWGAN-GP | TabDDPM | Ours |
|---|---|---|---|---|---|---|---|---|---|---|
| abalone9-18 | 0.6825 | 0.7681 | 0.7739 | 0.7823 | 0.7738 | 0.7659 | 0.7045 | 0.7472 | 0.7529 | **0.7837** |
| ecoli4 | 0.8341 | 0.8933 | 0.9001 | 0.9002 | 0.9099 | 0.9050 | 0.8368 | 0.9024 | 0.9171 | 0.9138 |
| page-blocks-1-3-vs-4 | 0.9287 | 0.9595 | 0.9499 | 0.9604 | 0.9635 | 0.9618 | 0.9153 | 0.9533 | 0.9522 | 0.9565 |
| yeast5 | 0.8188 | 0.9127 | 0.9117 | 0.9220 | 0.9183 | 0.9061 | 0.8389 | 0.9115 | 0.9122 | **0.9235** |
| yeast-0-5-6-7-9-vs-4 | 0.7546 | 0.7948 | 0.7925 | 0.7974 | 0.8060 | 0.7891 | 0.7652 | 0.8053 | 0.8061 | **0.8116** |
| yeast-2-vs-4 | 0.7342 | 0.7622 | 0.7646 | 0.8901 | 0.8904 | 0.8855 | 0.8649 | 0.8941 | 0.8878 | **0.8962** |
| ecoli1 | 0.8779 | 0.8811 | 0.8748 | 0.8812 | 0.8763 | 0.8751 | 0.8753 | 0.8790 | 0.8768 | **0.9538** |
| ecoli2 | 0.8822 | 0.8812 | 0.8826 | 0.8887 | 0.8846 | 0.8832 | 0.8540 | 0.8808 | 0.8830 | **0.9711** |
| ecoli3 | 0.7891 | 0.8560 | 0.8536 | 0.8583 | 0.8625 | 0.8488 | 0.8004 | 0.8682 | 0.8652 | **0.9542** |
| ecoli-0-vs-1 | 0.9341 | 0.9346 | 0.9317 | 0.9346 | 0.9345 | 0.9361 | 0.9328 | 0.9356 | 0.9354 | **0.9982** |
| glass-0-1-2-3- | 0.9528 | 0.9550 | 0.9557 | 0.9581 | 0.9556 | 0.9508 | 0.9493 | 0.9583 | 0.9578 | **0.9604** |
| vs-4-5-6 | | | | | | | | | | |
| haberman | 0.6235 | 0.6192 | 0.6377 | 0.6281 | 0.6451 | 0.6359 | 0.5976 | 0.6436 | 0.6358 | **0.6501** |
| newthyroid2 | 0.9369 | 0.9374 | 0.9388 | 0.9409 | 0.9418 | 0.9362 | 0.9253 | 0.9382 | 0.9376 | **0.9436** |
| segment0 | 0.9457 | 0.9462 | 0.9349 | 0.9454 | 0.9473 | 0.9468 | 0.9120 | 0.9382 | 0.9331 | **0.9494** |
| vehicle2 | 0.9168 | 0.9181 | 0.9197 | 0.9189 | 0.9220 | 0.9190 | 0.8877 | 0.9136 | 0.9067 | **0.9287** |
| yeast3 | 0.8780 | 0.8971 | 0.9002 | **0.9026** | 0.8970 | 0.8930 | 0.8540 | 0.8950 | 0.8900 | 0.8976 |
| heart | 0.8715 | 0.8675 | 0.8666 | 0.8682 | 0.8694 | 0.8700 | 0.8644 | **0.8737** | 0.8714 | 0.8720 |
| ionosphere | 0.9312 | 0.9309 | 0.9337 | 0.9294 | 0.9288 | 0.9274 | 0.9276 | 0.9286 | 0.9301 | **0.9344** |
| spambase | **0.9517** | 0.9501 | 0.9504 | 0.9489 | 0.9448 | 0.9503 | 0.9324 | 0.9449 | 0.9259 | 0.9259 |
| titanic | 0.7344 | 0.7353 | 0.7303 | 0.7341 | 0.7336 | **0.7356** | 0.7283 | 0.7312 | 0.7336 | 0.7327 |

TABLE VI

Mean and ranking of F1, G-mean, AUC for multiple oversampling algorithms over 20 datasets (higher ranking is better)

| | Average (F1) | average (G-mean) | Average (AUC) | Mean ranks (F1) | Mean ranks (G-mean) | Mean ranks (AUC) |
|---|---|---|---|---|---|---|
| SMOTE | 0.6565 | 0.7739 | 0.8700 | 5.40 | 5.30 | 4.60 |
| ADASYN | 0.6480 | 0.7707 | 0.8702 | 4.80 | 5.30 | 4.30 |
| OREM | 0.6763 | 0.7899 | 0.8795 | 6.35 | 6.30 | 6.05 |
| GDO | 0.6758 | 0.7862 | 0.8803 | 5.90 | 5.55 | 6.10 |
| RBO | 0.6650 | 0.7819 | 0.8761 | 4.95 | 5.40 | 4.65 |
| CGAN-GP | 0.5874 | 0.6623 | 0.8483 | 1.25 | 1.35 | 1.40 |
| CWAGN-GP | 0.6598 | 0.7652 | 0.8771 | 4.05 | 4.00 | 5.10 |
| TabDDPM | 0.6550 | 0.7703 | 0.8755 | 4.40 | 4.50 | 4.85 |
| Ours | **0.7059** | **0.8173** | **0.8979** | **7.90** | **7.30** | **7.95** |